\pdfoutput=1

\documentclass[11pt]{article}

\usepackage[final]{acl}

\usepackage{times}
\usepackage{latexsym}

\usepackage{listings}
\usepackage{url}
\usepackage{CJKutf8}

\usepackage[T1]{fontenc}

\usepackage[utf8]{inputenc}

\usepackage{microtype}

\definecolor{codegreen}{rgb}{0,0.6,0}
\definecolor{codegray}{rgb}{0.5,0.5,0.5}
\definecolor{codepurple}{rgb}{0.58,0,0.82}
\definecolor{backcolour}{rgb}{0.95,0.95,0.92}

\lstdefinestyle{mystyle}{
    backgroundcolor=\color{backcolour},   
    commentstyle=\color{codegreen},
    keywordstyle=\color{magenta},
    numberstyle=\tiny\color{codegray},
    stringstyle=\color{codepurple},
    basicstyle=\ttfamily\footnotesize,
    breakatwhitespace=false,         
    breaklines=true,                 
    captionpos=b,                    
    keepspaces=true,                 
    numbers=left,                    
    numbersep=5pt,                  
    showspaces=false,                
    showstringspaces=false,
    showtabs=false,                  
    tabsize=2
}

\usepackage{inconsolata}
\usepackage{times}
\usepackage{latexsym}
\usepackage{xspace}
\usepackage{amsmath}
\usepackage{multirow}
\usepackage{multicol}
\usepackage{pifont}
\usepackage{array}
\usepackage{arydshln}
\usepackage{booktabs}
\usepackage{graphicx}
\usepackage{lipsum}
\usepackage{soul} 
\usepackage{xcolor}

\usepackage{color-edits}
\addauthor{gn}{magenta}

\definecolor{cornflowerblue}{rgb}{0.39, 0.58, 0.93}
\definecolor{babypink}{rgb}{0.99, 0.26, 0.76}
\newcommand{\hlblue}[1]{\colorbox{cornflowerblue!30}{#1}}
\newcommand{\hlpink}[1]{\colorbox{babypink!40}{#1}}

\title{What Is Missing in Multilingual Visual Reasoning and How to Fix It}

\author{Yueqi Song, Simran Khanuja, Graham Neubig \\
Carnegie Mellon University\\}

\author{Yueqi Song, Simran Khanuja, Graham Neubig\\
\texttt{\{yueqis,gneubig\}@cs.cmu.edu}
\\[1em]
\makebox[\textwidth]{\fontsize{11}{11}\selectfont Carnegie Mellon University}
}

\begin{document}

\include{defs}
\maketitle
\begin{abstract}
\noindent NLP models today strive for supporting multiple languages and modalities, improving accessibility for diverse users. In this paper, we evaluate their multilingual, multimodal capabilities by testing on a visual reasoning task. We observe that proprietary systems like GPT-4V obtain the best performance on this task now, but open models lag in comparison. Surprisingly, GPT-4V exhibits similar performance between English and other languages, indicating the potential for equitable system development across languages. Our analysis on model failures reveals three key aspects that make this task challenging: \emph{multilinguality}, \emph{complex reasoning}, and \emph{multimodality}. To address these challenges, we propose three targeted interventions including a translate-test approach to tackle \emph{multilinguality}, a visual programming approach to break down \emph{complex reasoning}, and a method that leverages image captioning to address \emph{multimodality}. Our interventions achieve the \emph{best} open performance on this task in a \emph{zero-shot} setting, boosting open models LLaVA-v1.5-13B by 13.4\%, LLaVA-v1.6-34B by 20.3\%, and Qwen-VL by 16.7\%, while also minorly improving GPT-4V's performance.\footnote{The code implementations and prompts can be found at \url{https://github.com/yueqis/Multilingual_Visual_Reasoning}.}

\end{abstract}



\section{Introduction}


Language technology today is continually evolving to be more inclusive, with models becoming increasing multilingual \cite{lai2023chatgpt, li2022pretrained}, multimodal \cite{yang2023dawn}, or both \citep{chen2020uniter, zeng-etal-2023-cross, geigle2023mblip, achiam2023gpt}.
Even though this promotes broader user accessibility, past research has consistently highlighted differences in model performance across languages \citep{blasi-etal-2022-systematic} and cultures \citep{liu-etal-2021-visually}. Notably, these models often favor North American or Western contexts, potentially leaving behind users from other regions. \citep{liu-etal-2021-visually, hershcovich-etal-2022-challenges}. 

\begin{figure}[t]
    \includegraphics[width=\columnwidth]{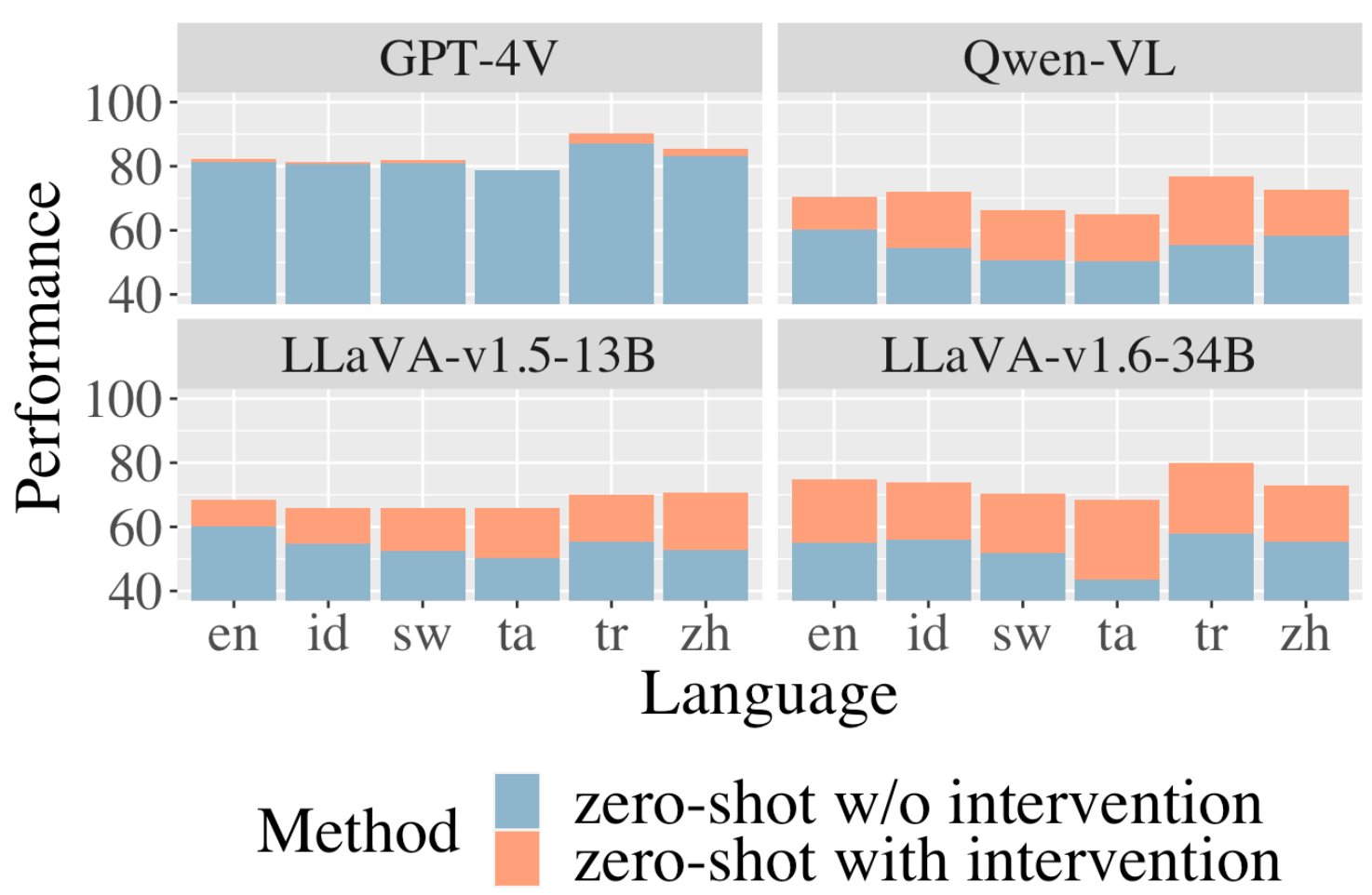}
    \caption{\emph{Our Contributions}: First, we evaluate the multilingual visual reasoning abilities of various models; then, we analyze key challenges where models are falling short; lastly, we propose three interventions to address these challenges.}
    \label{fig:main}
\end{figure}

The NLP community is currently witnessing a trend of moving away from openly releasing models to limiting their access through paid web APIs \cite{abdalla-etal-2023-elephant}. Additionally, the cost to use these services is often higher for low-resourced languages, despite poorer performance \cite{ahia-etal-2023-languages}. While it is certainly desirable to have strong and inclusive models available regardless of the access method, open, well-documented, and reasonably sized models have advantages from the point of view of control, ownership, cost, and advancing scientific understanding.

In this work, we first compare and contrast the multilingual, multicultural capabilities of the proprietary systems GPT-4V(ision) \cite{achiam2023gpt} and Gemini 1.5 Pro \cite{team2023gemini} with a plethora of open models like LLaVA \cite{liu2023llava, liu2023improvedllava, liu2024llavanext}, Qwen-VL \cite{Qwen-VL}, Qwen2-VL \cite{Qwen2-VL}, Cambrian \cite{tong2024cambrian}, Molmo \cite{deitke2024molmo}, Llama \cite{llamateam2024llama3herd}, mBLIP \cite{geigle2023mblip}, CCLM \cite{zeng-etal-2023-cross}, using two datasets on the same task of reasoning over texts and pairs of images, NLVR2 \citep{suhr-etal-2019-corpus} and MaRVL \cite{liu-etal-2021-visually}. We discuss this setup in more details in \S\ref{sec:background} and \S\ref{sec:protocol}. We find that GPT-4V significantly outperforms all open models. One additional unprecedented and surprising result is, as shown in Figure \ref{fig:main}, GPT-4V's consistency in performance across all languages, with performance on some even surpassing that on the NLVR2 dataset in English.
In contrast, as we will discuss in \S\ref{sec:results}, most open models still show a significant gap between English and other languages, perhaps due to deficiencies in training data, or due to the well-known ``curse of multilinguality'', where smaller models are less adept at processing low-resource languages \citep{conneau-etal-2020-unsupervised}.
This begs the question: ``how can we take open models, and bring them closer to achieving the exciting language-equitable multimodal reasoning results demonstrated by the opaque (and presumably bigger) GPT-4V?'' 


Towards this end, we conduct a careful analysis of the results from testing models on the multilingual visual reasoning task and discover that failures can arise from any of the three challenging aspects of the task: \emph{multilinguality}, \emph{reasoning}, and \emph{multimodality}. For \emph{multilinguality}, we find a significant gap in performance for other languages as compared to English. For \emph{reasoning}, we find a negative correlation of performance and the compositionality of the statement. For \emph{multimodality}, we find that models were typically pretrained on single image-text pairs, but haven't seen pairs of images in pretraining, which may lead to a mismatch between pretraining and evaluation objectives. We will discuss this in more details in \S\ref{sec:analysis}.

Based on our analysis, we design three interventions that address these challenges in section \ref{sec:intervention}. The first simply tackles \emph{multilinguality} -- we translate the MaRVL statements to English. Surprisingly, translation leads to a drop in performance for GPT-4V and Gemini-1.5-Pro (which might indicate their advanced multilingual capabilities), but helps improve performance for open models. Our next intervention tackles both \emph{multilinguality and reasoning}, by generating programs to reason over the set of images using the translated statements, inspired by \citet{gupta2023visual}'s VisProg method. Our third (and most effective) intervention tackles \emph{all three} challenges by first captioning images conditioned on the statement, and then reasoning over the captions, rather than the images, using chain-of-thought capabilities of text-modality LLMs \cite{wei2022chain}. Using this intervention, we obtain state-of-the-art zero-shot performance on the MaRVL dataset for open models, and also slightly improve the performance of GPT-4V itself, as shown in Figure \ref{fig:main}.

\section{Dataset Description}
\label{sec:background}

\begin{figure}
    \centering
    \includegraphics[width=\columnwidth]{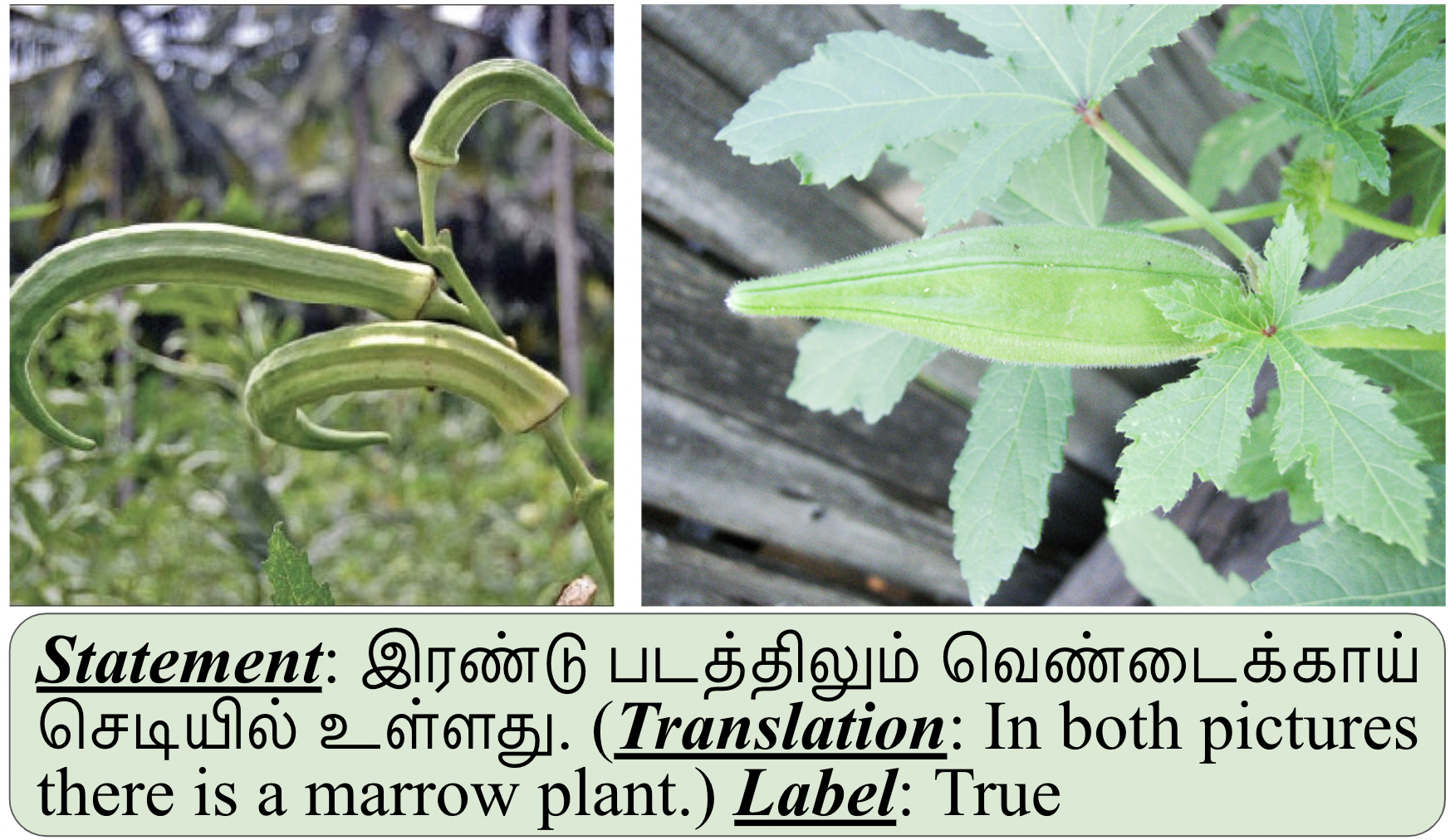}
    \caption{Example from the MaRVL Dataset: Given two images and a statement, the task is to infer whether the statement is true or false of the given image pair.}
    \label{fig:example}
\end{figure}

We evaluate on two visual reasoning datasets, both containing a statement in natural language and a pair of images. The task is to reason whether the statement is true based on the images, requiring reasoning over both images and the statement together.
Figure \ref{fig:example} shows an example of this task.

\paragraph{NLVR2}  NLVR2 contains 107,292 examples of English sentences with web photographs. Annotators paired visually-rich images and were encouraged to come up with compositional and linguistically diverse statements for each pair. The dataset contains a train-validation-test split. Images were collected using search queries generated from synsets derived from the ILSVRC2014 ImageNet challenge \cite{russakovsky2015imagenet}, with each query resulting in 4 pairs of images from Google Images\footnote{\url{https://images.google.com/}}. Queries for ImageNet \cite{deng2009imagenet} are based on the English WordNet \cite{poli2010theory}, whose concepts are more reflective of the North-American or Western cultures.

\paragraph{MaRVL} MaRVL explores the same task as NLVR2 in multilingual multicultural contexts. MaRVL is a test-only dataset collected for five diverse languages: Indonesian, Swahili, Tamil, Turkish, and Mandarin Chinese. Native speakers first select concepts that are reflective of their speaking population. Next, they curate images from the web that reflect those concepts within their specific cultural context. Finally, native speakers pair and write statements for each image pair, following the same protocol as that laid out for NLVR2.

\begin{table*}[!h]
    \resizebox{\linewidth}{!}{%
    \begin{tabular}{c c c c c c c c c}
\toprule
     \textbf{Model} & \textbf{NLVR2-en} & \textbf{id} & \textbf{sw} & \textbf{ta} & \textbf{tr} & \textbf{zh} & \textbf{MaRVL-Avg.} & \textbf{MaRVL-Avg. - EN}\\
\midrule
\hlblue{Human} & 96.2 & 96.3 & 93.0 & 98.0 & 97.0 & 95.5 & 96.0 & -0.2\\
\midrule
\multicolumn{9}{c}{\textit{Zero-Shot}}\\
\midrule
\hlpink{GPT-4V} & 81.4 & \textbf{80.6} & \textbf{81.0} & \textbf{78.6} & \textbf{87.1} & \textbf{83.2} & \textbf{82.1} & \textbf{0.7} \\ \midrule
\hlpink{Gemini 1.5 Pro} & 76.4 & 71.2 & 67.8 & 70.0 & 75.4 & 75.8 & 72.0 & -4.4\\ \midrule
mBLIP (mT0-XL)  & 67.3 & 64.9 & 64.8 & 69.6 & 68.0 & 65.9 & 66.6 & -0.7 \\
\midrule
LLaVA-v1.5-13B & 60.1 & 54.8 & 52.6 & 50.2 & 55.3 & 52.9 & 53.2 & -6.9 \\
\midrule
LLaVA-v1.6-34B & 54.9 & 56.0 & 51.8 & 43.4 & 57.9 & 55.3 & 52.9 & -2.0 \\
\midrule
Qwen-VL & 60.3 & 54.5 & 50.7 & 50.3 & 55.4 & 58.4 & 53.9 & -6.4 \\
\midrule
Qwen2-VL-7B-Instruct & 81.5 & 73.5 & 54.8 & 60.5 & 69.9 & 75.1 & 66.2 & -15.3 \\
\midrule
Cambrian-8B & 75.4 & 64.7 & 53.6 & 56.7 & 65.2 & 68.9 & 61.8 & -13.6 \\
\midrule
Molmo-7B & 65.3 & 61.1 & 49.6 & 49.6 & 52.2 & 62.2 & 54.9 & -10.4 \\
\midrule
Llama3.2-11B & 64.5 & 62.7 & 52.4 & 54.0 & 61.6 & 59.5 & 58.0 & -6.5 \\
\midrule
\multicolumn{9}{c}{\textit{Finetuned}}\\
\midrule
mBLIP (mT0-XL) & \textbf{85.2} & 75.1 & 74.6 & 75.9 & 74.3 & 75.7 & 75.1 & -10.1\\
\midrule
CCLM-4M & 80.2 & 67.6 & 64.4 & 60.5 & 69.0 & 69.2 & 66.1 & -14.1 \\
\midrule
xUNITER & 72.3 & 57.7 & 56.1 & 54.3 & 57.6 & 54.7 & 56.1 & -16.2\\
\midrule
mUNITER & 73.2 & 55.0 & 51.5 & 52.2 & 54.7 & 56.8 & 54.0 & -19.2 \\

\bottomrule
\end{tabular}}
\caption{NLVR2 and MaRVL performance across \hlblue{Human}, \hlpink{Proprietary Models}, and Open Models. Overall, mBLIP outperforms GPT-4V in NLVR2 post finetuning, while GPT-4V shows the best performance across all other languages without finetuning.}
\label{tab:original}    
\end{table*}






\section{Models and Evaluation Protocols}

\label{sec:protocol}

We evaluate various open models, including mBLIP (mt0-xl) \cite{geigle2023mblip}, LLaVA \cite{liu2023improvedllava, liu2024llavanext}, Qwen-VL \cite{Qwen-VL}, Qwen2-VL-7B-Instruct \cite{Qwen2-VL}, Cambrian-8B \cite{tong2024cambrian}, Molmo-7B \cite{deitke2024molmo}, CCLM \cite{zeng-etal-2023-cross}, and UNITERs \cite{chen2020uniter}; and a proprietary model GPT-4V(ision).\footnote{\textit{gpt-4-vision-preview} (\url{https://openai.com/research/gpt-4v-system-card}), abbreviated as "\textit{GPT-4V}".} We describe these models in \S\ref{sec:models}.
We evaluate them in two settings:

\textbf{Zero-shot.}
In this setting, models are not specifically fine-tuned for the task of visual reasoning.
This setting is academically interesting, as it more generally tests the ability of models to perform tasks, and the results are more likely to be representative of performance on datasets for which training data is not available.
In addition, it is practically useful since it can also be applied to LMs that cannot as easily be fine-tuned, such as the proprietary models GPT-4V and Gemini 1.5 Pro (due to their closed nature), and some large open models such as LLaVA and Qwen-VL (due to their relatively large sizes). We test LLaVA, Qwen-VL, Qwen2-VL-7B-Instruct, Cambrian-8B, Molmo-7B, mBLIP, GPT-4V, and Gemini-1.5-Pro in this setting.

\textbf{Finetuned.}
We finetune models that can more easily be finetuned on the English NLVR2 dataset, and test on NLVR2 and MaRVL.
This represents the realistic setting, adapting multilingual models to particular tasks using English data, which is relatively available.
We test mBLIP, CCLM-4M, xUNITER, and mUNITER in this setting.




\section{How well do proprietary and open models perform on multilingual visual reasoning?}
\label{sec:results}


In this section, we perform an examination of how-well these various models perform on multilingual multimodal reasoning tasks.
Table \ref{tab:original} shows performance of humans, open models, and proprietary models. For the models, we use the experiment protocols as in \S\ref{sec:protocol} in the zero-shot and finetuned settings. We ask the following questions:

\textbf{Which model performs the best?} \emph{Answer:} GPT-4V on MaRVL, and mBLIP (mT0-XL) on English post-fintuning. However, in the zero-shot setting, the proprietary model GPT-4V performs the best across all languages other than English,\footnote{We put GPT-4V in the zero-shot category because we evaluate the performance of GPT-4V on NLVR2 and MaRVL without finetuning on the NLVR2 training data. However, we do not know if GPT-4V has seen examples of NLVR2 or MaRVL during pretraining.} and open models lag behind especially in the multilingual setting. Note that despite GPT-4V's impressive performance, it still lags behind human performance by 10\% to 20\% across all languages, emphasizing that this task still is not completely solved.

\textbf{Which open model performs the best?} \emph{Answer:} mBLIP (mT0-XL), regardless of whether it is finetuned. The other open LMMs, for example LLaVA and Qwen-VL, are not explicitly trained on multilingual data, so the gap in MaRVL and NLVR2 performance is expected.

\textbf{Do models perform equitably across languages?}
Under zero-shot setting, GPT-4V and mBLIP both show equitable performance across languages, which is encouraging, although the latter significantly lags in overall performance compared to GPT-4V. Interestingly, post finetuning on NLVR2, mBLIP shows better performance on NLVR2 than GPT-4V, but still has lower performance on MaRVL. The gap between English and MaRVL languages also significantly increases for mBLIP from the zero-shot to finetuned setting. Maintaining the equity in performance across languages during finetuning is an interesting future direction, which may help models surpass GPT-4V's performance on multilingual visual reasoning. Other models lag mBLIP, both in overall performance and equity with English.

\section{What makes multilingual visual reasoning challenging?}

As noted in Table \ref{tab:original}, the best model still lags human performance by 10\% to 20\%. In this section, we aim to analyze what makes multilingual visual reasoning so challenging, and identify three key aspects as detailed below: 
\label{sec:analysis}
\subsection{Multilinguality and Sub-Optimal Cross-Lingual Transfer}
\label{sec:language-analysis}
In the finetuned setting, we observe a significant drop in performance for MaRVL languages as compared to NLVR2 in English. This is expected since models are finetuned only in English but not in these languages due to lack of training data.
We also note that performance on Swahili is consistently lower across models (excluding GPT-4V), which is the lowest-resource language amongst MaRVL languages, as laid out by the language resource taxonomy \cite{joshi-etal-2020-state}. This observation motivates us to evaluate models with MaRVL data translated to English, as we discuss in \S\ref{subsec:translate}.

In the zero-shot setting, GPT-4V and mBLIP both exhibit equitable performance on MaRVL as with NLVR2. Gemini 1.5 Pro also demonstrates equitable performance among languages to some extent. While LLaVA, Cambrian, Molmo, and Llama are not expected to perform as well for non-English languages and Qwen is not expected to perform as well for non-English and non-Chinese languages, they have poorer performance than mBLIP on NLVR2. While mBLIP is pretrained on multilingual multimodal data, LLaVA is not specifically trained on multilingual data. However, Qwen-VL is pretrained on Chinese data \cite{Qwen-VL}, and it is generally believed that LLaVA has multilingual abilities as it has seen multilingual data during pretraining \cite{liu2023llava, liu2023improvedllava, liu2024llavanext}. 

Overall, models have better visual reasoning abilities when given English inputs from US/European-centric cultures, while still lagging behind when facing multilingual and multicultural inputs.



\subsection{Complex Reasoning}

Data points in both NLVR2 and MaRVL require complex reasoning. An example statement from NLVR2 is "one image includes a silver stylus and a device with a blue keyboard base and an open screen propped up like an easel", which is semantically diverse, long in length, and has a compositional structure, requiring models to perform compositional and complex reasoning to infer the label.

\begin{figure}
    \centering
    \includegraphics[width=\columnwidth]{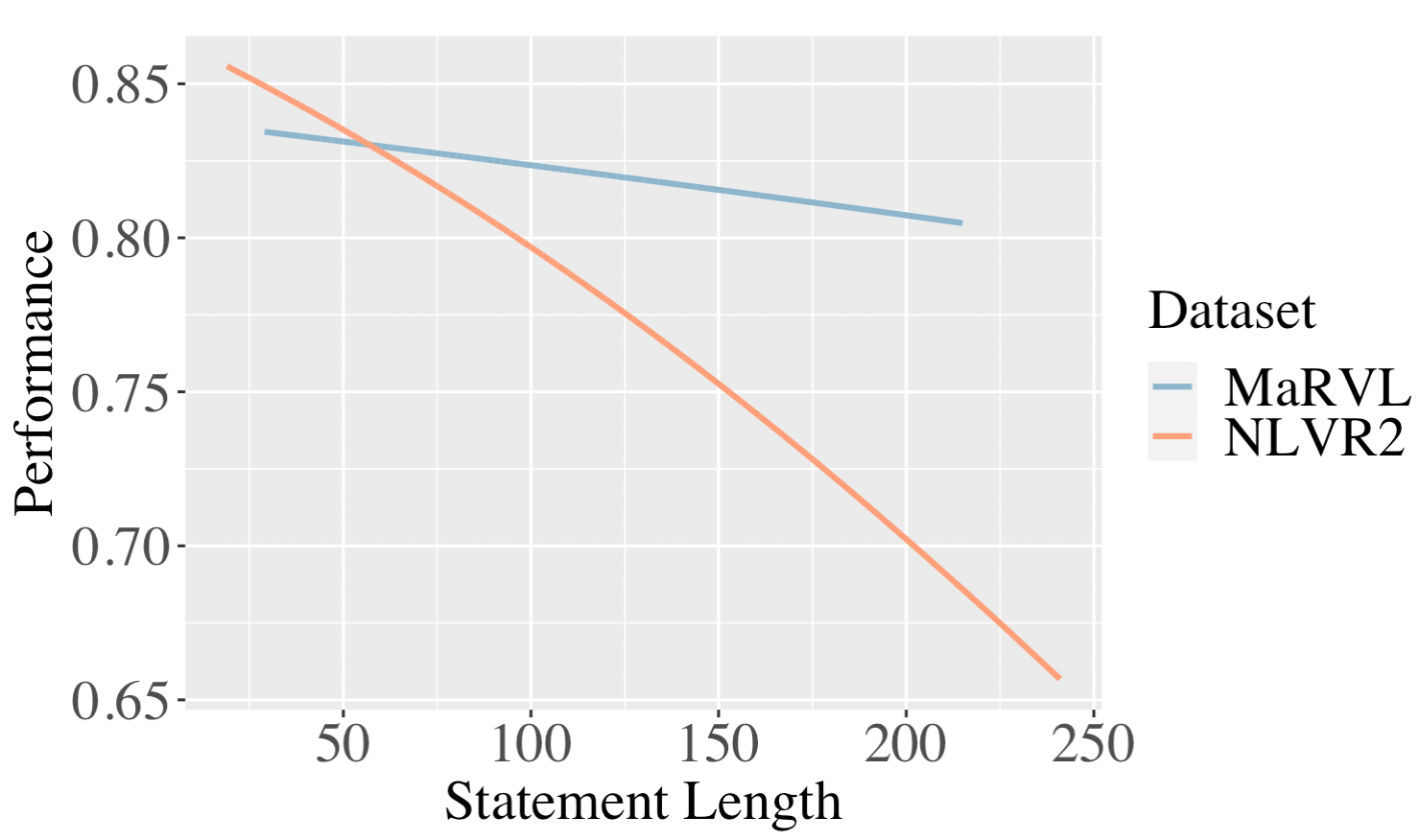}
    \caption{Performance of GPT-4V decreases as statement length increases.}
    \label{fig:length}
\end{figure}

As a proxy to the complexity of reasoning, we measure the number of words of the NLVR2 and MaRVL statements (translated to English), and find that model performances drop as the number of words of the statement increases. Figure \ref{fig:length} shows a graph of the performance of GPT-4V plotted against the number of words in each statement. We can clearly see a downward trend in the graph. Based on this, we are motivated to examine methods that break down long, compositional statements, as will be discussed in \S\ref{subsec:visprog}.

\begin{figure*}[!h]
    \includegraphics[width=\textwidth]{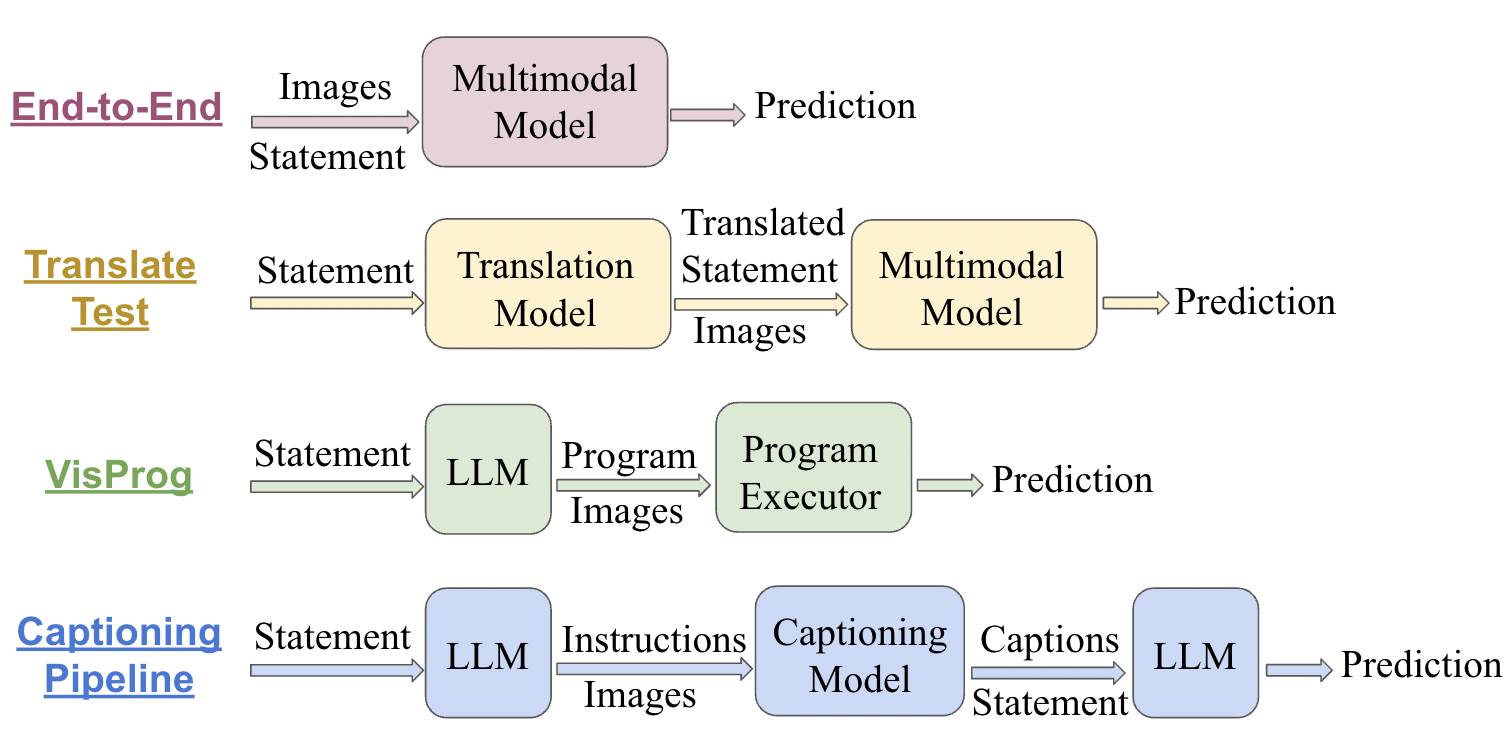}
    \caption{Flow chart visualizing the end-to-end testing in \S\ref{sec:results} and all interventions performed in \S\ref{sec:intervention}.}
    \label{fig:intervention}
\end{figure*}

\begin{table*}[!h]
    \resizebox{\linewidth}{!}{%
    \begin{tabular}{c c c c c c c c c}
    \toprule
         \textbf{Model} & \textbf{NLVR2-en} & \textbf{id} & \textbf{sw} & \textbf{ta} & \textbf{tr} & \textbf{zh} & \textbf{MaRVL-Avg.} & \textbf{MaRVL-Avg. - EN}\\
         \midrule
         \multicolumn{8}{c}{\textit{Zero-Shot}}\\
        \midrule
        \hlpink{GPT-4V} & 81.4 & 78.4 & 75.5 & 70.2 & 78.2 & 78.4 & 76.1 & -5.3\\
        \midrule
        \hlpink{Gemini 1.5 Pro} & 76.4 & 70.1 & 65.3 & 71.0 & 71.9 & 72.6 & 70.2 & -6.2\\
        \midrule
        \hlblue{LLaVA-v1.5-13B} & 60.1 & 53.1 & 53.9 & 54.1 & 58.3 & 54.0 & 54.7 & -5.4 \\
        \midrule
        \hlblue{LLaVA-v1.6-34B} & 54.9 & 55.7 & 53.1 & 52.8 & 55.3 & 55.4 & 54.5 & -0.4 \\
        \midrule
        \hlblue{Qwen-VL} & 60.3 & 58.2 & 56.0 & 58.8 & 63.0 & 58.4 & 58.9 & -1.42 \\
        \midrule
        \multicolumn{8}{c}{\textit{Finetuned}}\\
        \midrule
        \hlblue{CCLM-4M} & 80.2 & 72.3 & 69.2 & 69.7 & 77.6 & 71.8 & 72.1 & -8.1\\
        \midrule
        \hlblue{xUNITER} & 72.3 & 63.2 & 63.8 & 62.1 & 67.5 & 62.1 & 63.7 & -8.6\\
       \midrule
        \hlblue{mUNITER} & 73.2 & 59.8 & 63.4 & 62.3 & 69.2 & 62.7 & 63.5 & -9.7\\
        \midrule
        \hlblue{ViLT} & 73.7 & 61.7 & 62.0 & 65.1 & 69.8 & 60.9 & 63.9 & 9.8\\
    \bottomrule
    \end{tabular}}
\caption{MaRVL translate-test accuracies across \hlblue{Open} and \hlpink{Proprietary} models.}
\label{tab:translate}
\end{table*}

\subsection{Multimodality and Mismatch between Pretraining \& Evaluation}

NLVR2 and MaRVL contain two images per instance, along with a statement describing them, while vision-language models are typically trained on a single image-text pair \cite{cao2020behind}, leading to a mismatch in the input between pretraining and evaluation. Further, multimodal reasoning is known to be harder than reasoning over text alone \cite{mogadala2021trends, park2023visual}. Although Qwen has seen multi-image inputs during training \cite{Qwen-VL}, it still encounters difficulties in handling the complexities presented by multimodal reasoning during evaluation.

These, and the inherent difficulty of aligning image data and text data during the reasoning process make this task particularly challenging. This motivates us to (1) move from processing a pair of images together to processing each image separately; and (2) break down the two modalities of image and text in the reasoning process, as in \S\ref{subsec:captioning}.

\section{How can we address these challenges?}
\label{sec:intervention}

Based on our analysis from the previous section, we now move on to examining whether we can devise methods to further improve multilingual multimodal reasoning abilities, particularly those of open models. We examine three research questions, which we discuss in more details in the following subsections respectively. We will focus on a subset of the models from Section \ref{sec:protocol}. Figure \ref{fig:intervention} shows a flow chart visualizing the interventions we perform to address the research questions\footnote{\S\ref{sec:cost} discusses additional computation cost incurred by the interventions.}. 

\noindent \textbf{RQ1) (\emph{multilinguality})}  Does translating the text to English and reducing the cross-lingual gap aid performance? \emph{\textbf{Short Answer}}: it depends. 

\noindent \textbf{RQ2) (\emph{multilinguality+reasoning})} Can we break down the complex reasoning into modular programs which can be executed on a vision-text input? \emph{\textbf{Short Answer}}: yes, we adopt the Visual Programming approach \cite{gupta2023visual}.

\noindent \textbf{RQ3) (\emph{multilinguality+reasoning+multimodality})} Can we alleviate the need for multimodal interaction during the reasoning process? \emph{\textbf{Short Answer}}: yes, we propose a new approach utilizing captions.

\subsection{\emph{Addressing Multilinguality:} Translate-Test}
\label{subsec:translate}

In \S\ref{sec:language-analysis}, we find performance on NLVR2 is much better than MaRVL. While both are visual reasoning datasets, MaRVL is multi-cultural and contains data in 5 diverse languages. Since NLP systems perform significantly better with English data \cite{song-etal-2023-globalbench}, we first simply translate the reasoning statements to English using the Google Translate API \cite{wu2016google}. A visualization of the process of translate test can be found in Figure \ref{fig:intervention}.

In addition to the models we evaluate in \S\ref{sec:protocol}, we also evaluate ViLT \cite{kim2021vilt} for better comparisons, as our next proposed intervention in \S\ref{subsec:visprog} uses ViLT. We didn't evaluate ViLT on MaRVL before translate test, since it doesn't support the MaRVL languages. Our evaluation protocols follows the ones introduced in \S\ref{sec:protocol} and results are shown in Table \ref{tab:translate}. 

All prior works, as per our knowledge, have observed a gain in performance post translating to English \cite{liu-etal-2021-visually}. Our observation is consistent with prior findings for all models, except GPT-4V(ision) and Gemini 1.5 Pro. All models except for GPT-4V and Gemini 1.5 Pro see an increase in accuracy after performing translate test; while surprisingly, GPT-4V and Gemini 1.5 Pro show a sharp decrease in performance across almost all MaRVL languages after translate test. However, this is encouraging, because it speaks for the multilingual capabilities of these models, and indicates that the gains provided by translating to English are lower than the errors made in translating cultural-specific nuances in meaning. 

\begin{CJK*}{UTF8}{gbsn}
For example, the MaRVL statement "右图有青绿色的苹果" is translated to "the picture on the right has turquoise apples", where "青绿色" is translated to "turquoise". However, the color "青绿色" means pure green with a little bit cyan in Mandarin Chinese, which is different from "turquoise". Given this, GPT-4V reasons correctly when provided the statement in Mandarin, but makes mistakes when given the translated statement\footnote{See discussion on whether translate test may introduce biases due to inaccuracies in translation in Appendix \ref{appendix:translation}.}.
\end{CJK*}

\subsection{Addressing \emph{Multilinguality + Reasoning:} Visual Programming}
\label{subsec:visprog}

To improve performance of LLMs on reasoning tasks, beyond naive prompting, several methods have been introduced \cite{nye2021show, zhou2022least, wei2022chain, gao2023pal}. Particularly, PAL \cite{gao2023pal} provides significant improvements by decomposing a natural language instruction into multiple programmatic sub-modules, executed in an inference step to obtain the final answer. Most recently, efforts like VisProg \cite{gupta2023visual}, ViperGPT \cite{surismenon2023vipergpt}, Visual ChatGPT \cite{wu2023visual} have followed suit to solve multimodal reasoning using LLMs to generate \emph{visual} programs, that leverage off-the-shelf computer vision models for image processing during inference. Hence, we use VisProg to generate visual programs given translated statements as obtained in \S\ref{subsec:translate}. VisProg uses ViLT \cite{kim2021vilt} as its inherent vision module. 

Figure \ref{fig:intervention} shows the flow of VisProg. For example, given the statement: \emph{There is no one in the bedroom on the left, and there is someone in the bedroom on the right}, the generated visual program is:

\begin{lstlisting}[language=Python, caption=Visual program example, breaklines=true]
ANSWER0=VQA(image=LEFT,question='Is there anyone in the bedroom?')
ANSWER1=VQA(image=RIGHT,question='Is there anyone in the bedroom?')
ANSWER2=EVAL(ANSWER0 == False and ANSWER1 == True)
FINAL_ANSWER=RESULT(var=ANSWER2)
\end{lstlisting}

\begin{figure}[!h]
    \centering
    \includegraphics[width=\columnwidth]{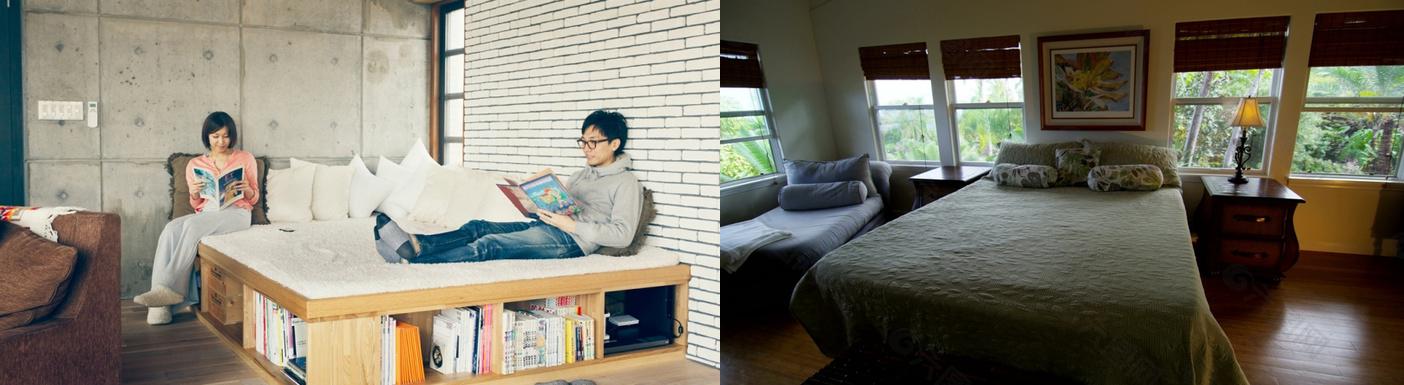}
    \caption{VisProg example image pair.}
    \label{fig:pal}
\end{figure}

If this program is executed on the images in Figure \ref{fig:pal}, then it will have ANSWER0 $=True$, ANSWER1 $=False$, so the final result is $False$.

For this intervention, we use text-davinci-003\footnote{text-davinci-003 is the model that the VisProg authors utilized when running VisProg.} as a representative of proprietary LLMs and LLaMA2-70B \cite{touvron2023llama} to represent open LLMs. Table \ref{tab:PAL} shows results to this method. Although this method does not achieve as high accuracy as models evaluated end-to-end in Table \ref{tab:original}, this approach provides valuable insights of breaking down complex reasoning into modular modules and utilizing prompts to make use of LLMs' strong in-context abilities. In addition, this approach, without any additional training, achieves on par performance on MaRVL, as compared to ViLT post-fintuning.

\begin{table}[!h]
    \resizebox{\columnwidth}{!}{%
    \begin{tabular}{c c c c c c c c}
    \toprule
         \multirow{2}{*}{\textbf{Model}} & \multirow{2}{*}{\textbf{NLVR}} & \multicolumn{6}{c}{\textbf{MaRVL}}\\
         \cdashline{3-8} \\ [-2.0ex]
         & & \textbf{id} & \textbf{sw} & \textbf{ta} & \textbf{tr} & \textbf{zh} & \textbf{Avg.}\\
    \midrule
        GPT-3 & 67.0 & 64.5 & 59.8 & 60.3 & 67.3 & 64.3 & 63.2\\
    \midrule
        LLaMA2-70b & 67.3 & 58.2 & 57.2 & 58.1 & 65.8 & 61.9 & 60.2\\
    \bottomrule
    \end{tabular}}
\caption{VisProg performance across models.}
\label{tab:PAL}    
\end{table}

\begin{table*}[hbt!]
    \resizebox{\linewidth}{!}{%
    \begin{tabular}{c c c c c c c c c}
    \toprule
         \textbf{Captioning} & \textbf{Reasoning} & \textbf{NLVR (en)}
         & \textbf{id} & \textbf{sw} & \textbf{ta} & \textbf{tr} & \textbf{zh} & \textbf{MaRVL-Avg.}\\
    \midrule
        InstructBLIP & LLaMA2-70B & 65.1 & 61.3 & 60.8 & 60.2 & 62.6 & 62.8 & 61.5\\
    \midrule
        PromptCap & LLaMA2-70B & 63.2 & 59.3 & 58.9 & 58.3 & 59.2 & 59.9 & 59.1\\
        \midrule
        \multirow{2}{*}{GPT-4V} & No Intervention & 81.4 & 80.6 & 81.0 & 78.6 & 87.1 & 83.2 & 82.1\\
         & GPT4 & 82.2 & 81.2 & 81.8 & 76.1 & 90.1 & 85.4 & 82.92\\
        \midrule
        \multirow{2}{*}{LLaVA-v1.5-13B} & No Intervention & 60.1 & 54.9 & 52.6 & 50.2 & 55.3 & 52.9 & 53.2 \\
        & LLaMA2-70B & 68.6 & 65.8 & 65.9 & 65.8 & 69.9 & 70.8 & 67.6\\
        \midrule
        \multirow{2}{*}{LLaVA-v1.6-34B} & No Intervention & 54.9 & 56.0 & 51.8 & 43.4 & 57.9 & 55.3 & 52.9 \\
        & LLaMA2-70B & 77.8 & 75.9 & 71.3 & 71.2 & 80.6 & 78.3 & 75.5 \\
        \midrule
        \multirow{2}{*}{Qwen-VL} & No Intervention & 60.3 & 54.5 & 50.7 & 50.3 & 55.4 & 58.4 & 53.9 \\
        & LLaMA2-70B & 70.3 & 72.1 & 66.3 & 65.1 & 76.7 & 72.8 & 70.6 \\
    \bottomrule
    \end{tabular}}
\caption{Captioning Pipeline Performance across Models. For rows with "No Intervention" stated in the "Reasoning" column, we pull over the end-to-end results of that model from Table 1, for the sake of comparison.}
\label{tab:question}    
\end{table*}

\subsection{Addressing \emph{Multilinguality + Reasoning + Multimodality}: Reasoning with Captions}

\label{subsec:captioning}

When analyzing errors for NLVR2, \citet{gupta2023visual} note that 69\% of them are caused by the vision module. This might be potentially worse for MaRVL, because open visual modules used in VisProg \cite{kim2021vilt} are trained on Western-centric datasets like Imagenet \cite{russakovsky2015imagenet}. Text-based LLMs, on the other hand, are trained on trillions of tokens, and are known to exhibit cultural awareness, albeit it may be limited \cite{yao2023empowering}. Hence, here we target the last remaining challenge, that of multimodal interaction needed for the reasoning process, by working with image captions instead of images. Concretely, we first caption both images, and use LLMs to reason about the statement with the two captions, instead of with the two images. Figure \ref{fig:intervention} shows a flow chart of how this pipeline works.


To make sure the captions capture necessary information needed for reasoning about the statement, as a first step of this intervention we use LLMs to generate targeted instructions based on the statement. Consider the statement "\textit{The picture on the left has several pencils of different colors, and the picture on the right has only one pencil}" from MaRVL-zh, the targeted instructions are:\\
\noindent\textbf{Left image} - "\textit{Write a short caption describing the number and colors of pencils};"\\
\textbf{Right image} - "\textit{Write a short caption describing the number of pencils}". 

\begin{figure}[!h]
    \centering
    \includegraphics[width=\columnwidth]{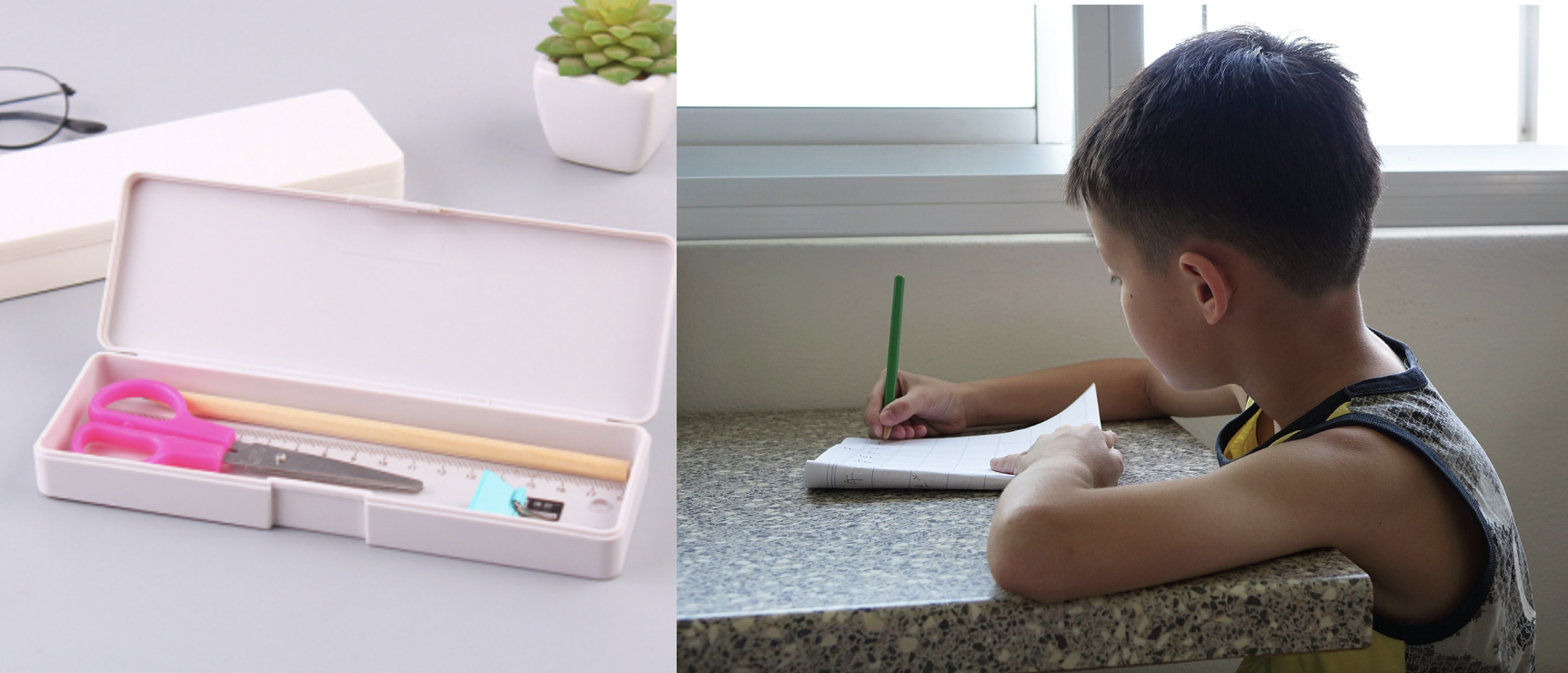}
    \caption{Captioning example image pair.}
    \label{fig:caption}
\end{figure}

As a second step, we generate captions following the targeted instructions in step 1, using various captioning models, including InstructBLIP \cite{liu2023improved}, PromptCap \cite{hu2022promptcap}, GPT-4V, LLaVA-v1.5-13B \cite{liu2023improvedllava}, LLaVA-v1.6-34B \cite{liu2024llavanext}, and Qwen-VL \cite{Qwen-VL}. The instructions can point them to focus on targeted contents in the image. For instance, for the statement in step 1 and the images in Figure \ref{fig:caption}, the captions generated by GPT-4V are: \\
\noindent\textbf{Left image} - \textit{A pencil case containing a single natural wood colored pencil.}; \\
\textbf{Right image} - \textit{A young mind sketches ideas with a classic green pencil.}.

Lastly, we prompt LLMs to reason whether the statement match caption pairs. For instance, for the example above, GPT-4 reasons as follows:\\
\noindent\textbf{Reasoning} - \textit{The statement is False. This is because the left image is described to include only a single pencil, which implies there are not several pencils of different colors in the left image. However, the statement claims that there should be several pencils of different colors in the left image. Meanwhile, the right image correctly shows a single green pencil, but the statement is negated by the inaccuracy of the part regarding the left image.}

Using this approach, visio-linguistic models only need to process one image at a time, instead of processing a pair of images simultaneously. In addition, reasoning is performed only over text, instead of over both image and text modalities.

Table \ref{tab:question} shows the performance of this intervention. Performance of InstructBLIP and PromptCap under zero-shot setting is on par with UNITER models post-finetuning. This intervention  improves performance of LLaVA-v1.5-13B by 10\% to 16\%, LLaVA-v1.6-34B by 18\% to 25\%, and Qwen-VL\footnote{\S\ref{sec:additional} discusses additional experiments on Qwen-VL.} by 10\% to 21\% depending on language, while also minorly improving GPT-4V's performance. On average, our pipeline boosts LLaVA-v1.5-13B's performance on MaRVL by 13.4\%, LLaVA-v1.6-34B's performance by 20.3\%, and Qwen-VL's performance by 16.7\%. This intervention improves performance of LLaVA and Qwen-VL, achieving the best performance under zero-shot setting (without training on reasoning of pairs of images). 

\section{Related Work}
\label{sec:related_work}


\paragraph{From Pretraining to Instruction Tuning}

Previous research on instruction tuning sparks multiple works to finetune models on instructions, and create general-purpose models that are good at performing tasks under zero-shot settings \cite{ouyang2022training, liu2023improved, geigle2023mblip}. However, instruction tuning data is mostly in English \cite{touvron2023llama, liu2023improved}. Due to the absence of multilingual instruction tuning data, models may struggle to effectively process multilingual inputs.

\paragraph{Moving Beyond English}

Past research efforts has predominantly centered around English language models, highlighting differences in model performance across languages  \citep{blasi-etal-2022-systematic, song-etal-2023-globalbench}. In the visio-linguistic domain, research in instruction tuning also center on English, due to a lack of multilingual instruction training data \cite{geigle2023mblip}. To this end, mBLIP \cite{geigle2023mblip} translated instruction training data to various languages, and perform instruction tuning. This is the first multilingual instruction tuned vision LLM.

\paragraph{Gap between Proprietary Models and Open Models} 

Currently, there is a trend of shifting from openly releasing models to paid APIs \cite{abdalla-etal-2023-elephant}. Previous research on examining GPT-4V and Gemini 1.5 Pro demonstrates its unprecedented multimodal capabilities, and there is still a gap between this proprietary model and other open source models \cite{yang2023dawn}. However, it is still important for the community to have as strong open source multimodal models.

\section{Conclusion}

In conclusion, we explore the evolving domain of multilingual visual reasoning. We observe a trend towards inclusivity in models, yet recognize persistent disparities in performance across languages and cultures. While proprietary systems like GPT-4V exhibit notable and equitable accuracy across languages, open models still face challenges in bridging the gap, especially for low-resource languages. Our analysis highlights the superior performance of GPT-4V but also underscores the need for advancements in open models. Leveraging interventions addressing multilinguality, multimodality, and reasoning, we demonstrate significant enhancements in open model performance, achieving state-of-the-art results under zero-shot settings for open models. Our findings emphasizes the potential for further advancements in multilingual visual reasoning, with the aim of narrowing down the gap between human and machine performance, and the gap between proprietary and open models.


\section*{Limitations}

With the goal of evaluating the multilingual visual reasoning capabilities of models, we employ NLVR2 and MaRVL, both of which engage in the task of determining whether a pair of images correspond to a given statement. This choice stems from MaRVL being the sole visual reasoning dataset with multilingual support, as far as our current knowledge extends. 

\paragraph{Representing Visual Reasoning} It's important to acknowledge that the task of NLVR2 and MaRVL solely represents a specific task of visual reasoning. Other aspects and dimensions of this domain may not be fully represented by this particular task.

\paragraph{Representing Multilinguality} In addition, note that the combination of NLVR2 and MaRVL covers 6 distinct languages: English, Indonesian, Swahili, Tamil, Turkish, and Mandarin Chinese. This is only a small subset of all languages worldwide.

\bibliography{anthology,custom}

\begin{thebibliography}{53}
\expandafter\ifx\csname natexlab\endcsname\relax\def\natexlab#1{#1}\fi

\bibitem[{Abdalla et~al.(2023)Abdalla, Wahle, Ruas, N{\'e}v{\'e}ol, Ducel, Mohammad, and Fort}]{abdalla-etal-2023-elephant}
Mohamed Abdalla, Jan~Philip Wahle, Terry Ruas, Aur{\'e}lie N{\'e}v{\'e}ol, Fanny Ducel, Saif Mohammad, and Karen Fort. 2023.
\newblock \href {https://doi.org/10.18653/v1/2023.acl-long.734} {The elephant in the room: Analyzing the presence of big tech in natural language processing research}.
\newblock In \emph{Proceedings of the 61st Annual Meeting of the Association for Computational Linguistics (Volume 1: Long Papers)}, pages 13141--13160, Toronto, Canada. Association for Computational Linguistics.

\bibitem[{Achiam et~al.(2023)Achiam, Adler, Agarwal, Ahmad, Akkaya, Aleman, Almeida, Altenschmidt, Altman, Anadkat et~al.}]{achiam2023gpt}
Josh Achiam, Steven Adler, Sandhini Agarwal, Lama Ahmad, Ilge Akkaya, Florencia~Leoni Aleman, Diogo Almeida, Janko Altenschmidt, Sam Altman, Shyamal Anadkat, et~al. 2023.
\newblock Gpt-4 technical report.
\newblock \emph{arXiv preprint arXiv:2303.08774}.

\bibitem[{Ahia et~al.(2023)Ahia, Kumar, Gonen, Kasai, Mortensen, Smith, and Tsvetkov}]{ahia-etal-2023-languages}
Orevaoghene Ahia, Sachin Kumar, Hila Gonen, Jungo Kasai, David Mortensen, Noah Smith, and Yulia Tsvetkov. 2023.
\newblock \href {https://doi.org/10.18653/v1/2023.emnlp-main.614} {Do all languages cost the same? tokenization in the era of commercial language models}.
\newblock In \emph{Proceedings of the 2023 Conference on Empirical Methods in Natural Language Processing}, pages 9904--9923, Singapore. Association for Computational Linguistics.

\bibitem[{Bai et~al.(2023{\natexlab{a}})Bai, Bai, Chu, Cui, Dang, Deng, Fan, Ge, Han, Huang, Hui, Ji, Li, Lin, Lin, Liu, Liu, Lu, Lu, Ma, Men, Ren, Ren, Tan, Tan, Tu, Wang, Wang, Wang, Wu, Xu, Xu, Yang, Yang, Yang, Yang, Yao, Yu, Yuan, Yuan, Zhang, Zhang, Zhang, Zhang, Zhou, Zhou, Zhou, and Zhu}]{qwen}
Jinze Bai, Shuai Bai, Yunfei Chu, Zeyu Cui, Kai Dang, Xiaodong Deng, Yang Fan, Wenbin Ge, Yu~Han, Fei Huang, Binyuan Hui, Luo Ji, Mei Li, Junyang Lin, Runji Lin, Dayiheng Liu, Gao Liu, Chengqiang Lu, Keming Lu, Jianxin Ma, Rui Men, Xingzhang Ren, Xuancheng Ren, Chuanqi Tan, Sinan Tan, Jianhong Tu, Peng Wang, Shijie Wang, Wei Wang, Shengguang Wu, Benfeng Xu, Jin Xu, An~Yang, Hao Yang, Jian Yang, Shusheng Yang, Yang Yao, Bowen Yu, Hongyi Yuan, Zheng Yuan, Jianwei Zhang, Xingxuan Zhang, Yichang Zhang, Zhenru Zhang, Chang Zhou, Jingren Zhou, Xiaohuan Zhou, and Tianhang Zhu. 2023{\natexlab{a}}.
\newblock Qwen technical report.
\newblock \emph{arXiv preprint arXiv:2309.16609}.

\bibitem[{Bai et~al.(2023{\natexlab{b}})Bai, Bai, Yang, Wang, Tan, Wang, Lin, Zhou, and Zhou}]{Qwen-VL}
Jinze Bai, Shuai Bai, Shusheng Yang, Shijie Wang, Sinan Tan, Peng Wang, Junyang Lin, Chang Zhou, and Jingren Zhou. 2023{\natexlab{b}}.
\newblock Qwen-vl: A versatile vision-language model for understanding, localization, text reading, and beyond.
\newblock \emph{arXiv preprint arXiv:2308.12966}.

\bibitem[{Blasi et~al.(2022)Blasi, Anastasopoulos, and Neubig}]{blasi-etal-2022-systematic}
Damian Blasi, Antonios Anastasopoulos, and Graham Neubig. 2022.
\newblock \href {https://doi.org/10.18653/v1/2022.acl-long.376} {Systematic inequalities in language technology performance across the world{'}s languages}.
\newblock In \emph{Proceedings of the 60th Annual Meeting of the Association for Computational Linguistics (Volume 1: Long Papers)}, pages 5486--5505, Dublin, Ireland. Association for Computational Linguistics.

\bibitem[{Cao et~al.(2020)Cao, Gan, Cheng, Yu, Chen, and Liu}]{cao2020behind}
Jize Cao, Zhe Gan, Yu~Cheng, Licheng Yu, Yen-Chun Chen, and Jingjing Liu. 2020.
\newblock Behind the scene: Revealing the secrets of pre-trained vision-and-language models.
\newblock In \emph{Computer Vision--ECCV 2020: 16th European Conference, Glasgow, UK, August 23--28, 2020, Proceedings, Part VI 16}, pages 565--580. Springer.

\bibitem[{Chen et~al.(2020)Chen, Li, Yu, El~Kholy, Ahmed, Gan, Cheng, and Liu}]{chen2020uniter}
Yen-Chun Chen, Linjie Li, Licheng Yu, Ahmed El~Kholy, Faisal Ahmed, Zhe Gan, Yu~Cheng, and Jingjing Liu. 2020.
\newblock Uniter: Universal image-text representation learning.
\newblock In \emph{European conference on computer vision}, pages 104--120. Springer.

\bibitem[{Chiang et~al.(2023)Chiang, Li, Lin, Sheng, Wu, Zhang, Zheng, Zhuang, Zhuang, Gonzalez, Stoica, and Xing}]{vicuna2023}
Wei-Lin Chiang, Zhuohan Li, Zi~Lin, Ying Sheng, Zhanghao Wu, Hao Zhang, Lianmin Zheng, Siyuan Zhuang, Yonghao Zhuang, Joseph~E. Gonzalez, Ion Stoica, and Eric~P. Xing. 2023.
\newblock \href {https://lmsys.org/blog/2023-03-30-vicuna/} {Vicuna: An open-source chatbot impressing gpt-4 with 90\%* chatgpt quality}.

\bibitem[{Chung et~al.(2022)Chung, Hou, Longpre, Zoph, Tay, Fedus, Li, Wang, Dehghani, Brahma et~al.}]{chung2022scaling}
Hyung~Won Chung, Le~Hou, Shayne Longpre, Barret Zoph, Yi~Tay, William Fedus, Yunxuan Li, Xuezhi Wang, Mostafa Dehghani, Siddhartha Brahma, et~al. 2022.
\newblock Scaling instruction-finetuned language models.
\newblock \emph{arXiv preprint arXiv:2210.11416}.

\bibitem[{Conneau et~al.(2020)Conneau, Khandelwal, Goyal, Chaudhary, Wenzek, Guzm{\'a}n, Grave, Ott, Zettlemoyer, and Stoyanov}]{conneau-etal-2020-unsupervised}
Alexis Conneau, Kartikay Khandelwal, Naman Goyal, Vishrav Chaudhary, Guillaume Wenzek, Francisco Guzm{\'a}n, Edouard Grave, Myle Ott, Luke Zettlemoyer, and Veselin Stoyanov. 2020.
\newblock \href {https://doi.org/10.18653/v1/2020.acl-main.747} {Unsupervised cross-lingual representation learning at scale}.
\newblock In \emph{Proceedings of the 58th Annual Meeting of the Association for Computational Linguistics}, pages 8440--8451, Online. Association for Computational Linguistics.

\bibitem[{Deitke et~al.(2024)Deitke, Clark, Lee, Tripathi, Yang, Park, Salehi, Muennighoff, Lo, Soldaini et~al.}]{deitke2024molmo}
Matt Deitke, Christopher Clark, Sangho Lee, Rohun Tripathi, Yue Yang, Jae~Sung Park, Mohammadreza Salehi, Niklas Muennighoff, Kyle Lo, Luca Soldaini, et~al. 2024.
\newblock Molmo and pixmo: Open weights and open data for state-of-the-art multimodal models.
\newblock \emph{arXiv preprint arXiv:2409.17146}.

\bibitem[{Deng et~al.(2009)Deng, Dong, Socher, Li, Li, and Fei-Fei}]{deng2009imagenet}
Jia Deng, Wei Dong, Richard Socher, Li-Jia Li, Kai Li, and Li~Fei-Fei. 2009.
\newblock Imagenet: A large-scale hierarchical image database.
\newblock In \emph{2009 IEEE conference on computer vision and pattern recognition}, pages 248--255. Ieee.

\bibitem[{Devlin et~al.(2018)Devlin, Chang, Lee, and Toutanova}]{devlin2018bert}
Jacob Devlin, Ming-Wei Chang, Kenton Lee, and Kristina Toutanova. 2018.
\newblock Bert: Pre-training of deep bidirectional transformers for language understanding.
\newblock \emph{arXiv preprint arXiv:1810.04805}.

\bibitem[{Gao et~al.(2023)Gao, Madaan, Zhou, Alon, Liu, Yang, Callan, and Neubig}]{gao2023pal}
Luyu Gao, Aman Madaan, Shuyan Zhou, Uri Alon, Pengfei Liu, Yiming Yang, Jamie Callan, and Graham Neubig. 2023.
\newblock Pal: Program-aided language models.
\newblock In \emph{International Conference on Machine Learning}, pages 10764--10799. PMLR.

\bibitem[{Geigle et~al.(2023)Geigle, Jain, Timofte, and Glava\v{s}}]{geigle2023mblip}
Gregor Geigle, Abhay Jain, Radu Timofte, and Goran Glava\v{s}. 2023.
\newblock \href {http://arxiv.org/abs/2307.06930} {mblip: Efficient bootstrapping of multilingual vision-llms}.
\newblock \emph{arXiv}, abs/2307.06930.

\bibitem[{{Gemini Team, Google}(2024{\natexlab{a}})}]{geminiteam2024gemini15unlockingmultimodal}
{Gemini Team, Google}. 2024{\natexlab{a}}.
\newblock \href {http://arxiv.org/abs/2403.05530} {Gemini 1.5: Unlocking multimodal understanding across millions of tokens of context}.

\bibitem[{{Gemini Team, Google}(2024{\natexlab{b}})}]{geminiteam2024geminifamilyhighlycapable}
{Gemini Team, Google}. 2024{\natexlab{b}}.
\newblock \href {http://arxiv.org/abs/2312.11805} {Gemini: A family of highly capable multimodal models}.

\bibitem[{Gupta and Kembhavi(2023)}]{gupta2023visual}
Tanmay Gupta and Aniruddha Kembhavi. 2023.
\newblock Visual programming: Compositional visual reasoning without training.
\newblock In \emph{Proceedings of the IEEE/CVF Conference on Computer Vision and Pattern Recognition}, pages 14953--14962.

\bibitem[{Hershcovich et~al.(2022)Hershcovich, Frank, Lent, de~Lhoneux, Abdou, Brandl, Bugliarello, Cabello~Piqueras, Chalkidis, Cui, Fierro, Margatina, Rust, and S{\o}gaard}]{hershcovich-etal-2022-challenges}
Daniel Hershcovich, Stella Frank, Heather Lent, Miryam de~Lhoneux, Mostafa Abdou, Stephanie Brandl, Emanuele Bugliarello, Laura Cabello~Piqueras, Ilias Chalkidis, Ruixiang Cui, Constanza Fierro, Katerina Margatina, Phillip Rust, and Anders S{\o}gaard. 2022.
\newblock \href {https://doi.org/10.18653/v1/2022.acl-long.482} {Challenges and strategies in cross-cultural {NLP}}.
\newblock In \emph{Proceedings of the 60th Annual Meeting of the Association for Computational Linguistics (Volume 1: Long Papers)}, pages 6997--7013, Dublin, Ireland. Association for Computational Linguistics.

\bibitem[{Hu et~al.(2022)Hu, Hua, Yang, Shi, Smith, and Luo}]{hu2022promptcap}
Yushi Hu, Hang Hua, Zhengyuan Yang, Weijia Shi, Noah~A Smith, and Jiebo Luo. 2022.
\newblock Promptcap: Prompt-guided task-aware image captioning.
\newblock \emph{arXiv preprint arXiv:2211.09699}.

\bibitem[{Joshi et~al.(2020)Joshi, Santy, Budhiraja, Bali, and Choudhury}]{joshi-etal-2020-state}
Pratik Joshi, Sebastin Santy, Amar Budhiraja, Kalika Bali, and Monojit Choudhury. 2020.
\newblock \href {https://doi.org/10.18653/v1/2020.acl-main.560} {The state and fate of linguistic diversity and inclusion in the {NLP} world}.
\newblock In \emph{Proceedings of the 58th Annual Meeting of the Association for Computational Linguistics}, pages 6282--6293, Online. Association for Computational Linguistics.

\bibitem[{Kim et~al.(2021)Kim, Son, and Kim}]{kim2021vilt}
Wonjae Kim, Bokyung Son, and Ildoo Kim. 2021.
\newblock Vilt: Vision-and-language transformer without convolution or region supervision.
\newblock In \emph{International Conference on Machine Learning}, pages 5583--5594. PMLR.

\bibitem[{Lai et~al.(2023)Lai, Ngo, Veyseh, Man, Dernoncourt, Bui, and Nguyen}]{lai2023chatgpt}
Viet~Dac Lai, Nghia~Trung Ngo, Amir Pouran~Ben Veyseh, Hieu Man, Franck Dernoncourt, Trung Bui, and Thien~Huu Nguyen. 2023.
\newblock Chatgpt beyond english: Towards a comprehensive evaluation of large language models in multilingual learning.
\newblock \emph{arXiv preprint arXiv:2304.05613}.

\bibitem[{Li et~al.(2022)Li, Tang, Zhao, Nie, and Wen}]{li2022pretrained}
Junyi Li, Tianyi Tang, Wayne~Xin Zhao, Jian-Yun Nie, and Ji-Rong Wen. 2022.
\newblock Pretrained language models for text generation: A survey.
\newblock \emph{arXiv preprint arXiv:2201.05273}.

\bibitem[{Liu et~al.(2021)Liu, Bugliarello, Ponti, Reddy, Collier, and Elliott}]{liu-etal-2021-visually}
Fangyu Liu, Emanuele Bugliarello, Edoardo~Maria Ponti, Siva Reddy, Nigel Collier, and Desmond Elliott. 2021.
\newblock \href {https://doi.org/10.18653/v1/2021.emnlp-main.818} {Visually grounded reasoning across languages and cultures}.
\newblock In \emph{Proceedings of the 2021 Conference on Empirical Methods in Natural Language Processing}, pages 10467--10485, Online and Punta Cana, Dominican Republic. Association for Computational Linguistics.

\bibitem[{Liu et~al.(2023{\natexlab{a}})Liu, Li, Li, and Lee}]{liu2023improvedllava}
Haotian Liu, Chunyuan Li, Yuheng Li, and Yong~Jae Lee. 2023{\natexlab{a}}.
\newblock Improved baselines with visual instruction tuning.

\bibitem[{Liu et~al.(2023{\natexlab{b}})Liu, Li, Li, and Lee}]{liu2023improved}
Haotian Liu, Chunyuan Li, Yuheng Li, and Yong~Jae Lee. 2023{\natexlab{b}}.
\newblock Improved baselines with visual instruction tuning.
\newblock \emph{arXiv preprint arXiv:2310.03744}.

\bibitem[{Liu et~al.(2024)Liu, Li, Li, Li, Zhang, Shen, and Lee}]{liu2024llavanext}
Haotian Liu, Chunyuan Li, Yuheng Li, Bo~Li, Yuanhan Zhang, Sheng Shen, and Yong~Jae Lee. 2024.
\newblock \href {https://llava-vl.github.io/blog/2024-01-30-llava-next/} {Llava-next: Improved reasoning, ocr, and world knowledge}.

\bibitem[{Liu et~al.(2023{\natexlab{c}})Liu, Li, Wu, and Lee}]{liu2023llava}
Haotian Liu, Chunyuan Li, Qingyang Wu, and Yong~Jae Lee. 2023{\natexlab{c}}.
\newblock Visual instruction tuning.
\newblock In \emph{NeurIPS}.

\bibitem[{{Llama Team, Meta}(2024)}]{llamateam2024llama3herd}
{Llama Team, Meta}. 2024.
\newblock \href {https://llama.meta.com/} {The llama 3 herd of models}.
\newblock Llama 3 is a set of multilingual language models supporting coding, reasoning, and tool usage. The largest model features 405B parameters and a 128K token context window. It delivers comparable performance to GPT-4 across a variety of tasks.

\bibitem[{Mogadala et~al.(2021)Mogadala, Kalimuthu, and Klakow}]{mogadala2021trends}
Aditya Mogadala, Marimuthu Kalimuthu, and Dietrich Klakow. 2021.
\newblock Trends in integration of vision and language research: A survey of tasks, datasets, and methods.
\newblock \emph{Journal of Artificial Intelligence Research}, 71:1183--1317.

\bibitem[{Nye et~al.(2021)Nye, Andreassen, Gur-Ari, Michalewski, Austin, Bieber, Dohan, Lewkowycz, Bosma, Luan et~al.}]{nye2021show}
Maxwell Nye, Anders~Johan Andreassen, Guy Gur-Ari, Henryk Michalewski, Jacob Austin, David Bieber, David Dohan, Aitor Lewkowycz, Maarten Bosma, David Luan, et~al. 2021.
\newblock Show your work: Scratchpads for intermediate computation with language models.
\newblock \emph{arXiv preprint arXiv:2112.00114}.

\bibitem[{Ouyang et~al.(2022)Ouyang, Wu, Jiang, Almeida, Wainwright, Mishkin, Zhang, Agarwal, Slama, Ray et~al.}]{ouyang2022training}
Long Ouyang, Jeffrey Wu, Xu~Jiang, Diogo Almeida, Carroll Wainwright, Pamela Mishkin, Chong Zhang, Sandhini Agarwal, Katarina Slama, Alex Ray, et~al. 2022.
\newblock Training language models to follow instructions with human feedback.
\newblock \emph{Advances in Neural Information Processing Systems}, 35:27730--27744.

\bibitem[{Park and Kim(2023)}]{park2023visual}
Sang-Min Park and Young-Gab Kim. 2023.
\newblock Visual language integration: A survey and open challenges.
\newblock \emph{Computer Science Review}, 48:100548.

\bibitem[{Poli et~al.(2010)Poli, Healy, and Kameas}]{poli2010theory}
Roberto Poli, Michael Healy, and Achilles Kameas. 2010.
\newblock \emph{Theory and applications of ontology: Computer applications}.
\newblock Springer.

\bibitem[{Russakovsky et~al.(2015)Russakovsky, Deng, Su, Krause, Satheesh, Ma, Huang, Karpathy, Khosla, Bernstein et~al.}]{russakovsky2015imagenet}
Olga Russakovsky, Jia Deng, Hao Su, Jonathan Krause, Sanjeev Satheesh, Sean Ma, Zhiheng Huang, Andrej Karpathy, Aditya Khosla, Michael Bernstein, et~al. 2015.
\newblock Imagenet large scale visual recognition challenge.
\newblock \emph{International journal of computer vision}, 115:211--252.

\bibitem[{Shazeer et~al.(2017)Shazeer, Mirhoseini, Maziarz, Davis, Le, Hinton, and Dean}]{shazeer2017outrageouslylargeneuralnetworks}
Noam Shazeer, Azalia Mirhoseini, Krzysztof Maziarz, Andy Davis, Quoc Le, Geoffrey Hinton, and Jeff Dean. 2017.
\newblock \href {http://arxiv.org/abs/1701.06538} {Outrageously large neural networks: The sparsely-gated mixture-of-experts layer}.

\bibitem[{Song et~al.(2023)Song, Khanuja, Liu, Faisal, Ostapenko, Winata, Aji, Cahyawijaya, Tsvetkov, Anastasopoulos, and Neubig}]{song-etal-2023-globalbench}
Yueqi Song, Simran Khanuja, Pengfei Liu, Fahim Faisal, Alissa Ostapenko, Genta Winata, Alham Aji, Samuel Cahyawijaya, Yulia Tsvetkov, Antonios Anastasopoulos, and Graham Neubig. 2023.
\newblock \href {https://doi.org/10.18653/v1/2023.emnlp-main.875} {{G}lobal{B}ench: A benchmark for global progress in natural language processing}.
\newblock In \emph{Proceedings of the 2023 Conference on Empirical Methods in Natural Language Processing}, pages 14157--14171, Singapore. Association for Computational Linguistics.

\bibitem[{Suhr et~al.(2019)Suhr, Zhou, Zhang, Zhang, Bai, and Artzi}]{suhr-etal-2019-corpus}
Alane Suhr, Stephanie Zhou, Ally Zhang, Iris Zhang, Huajun Bai, and Yoav Artzi. 2019.
\newblock \href {https://doi.org/10.18653/v1/P19-1644} {A corpus for reasoning about natural language grounded in photographs}.
\newblock In \emph{Proceedings of the 57th Annual Meeting of the Association for Computational Linguistics}, pages 6418--6428, Florence, Italy. Association for Computational Linguistics.

\bibitem[{Sur\'is et~al.(2023)Sur\'is, Menon, and Vondrick}]{surismenon2023vipergpt}
D\'idac Sur\'is, Sachit Menon, and Carl Vondrick. 2023.
\newblock Vipergpt: Visual inference via python execution for reasoning.
\newblock \emph{Proceedings of IEEE International Conference on Computer Vision (ICCV)}.

\bibitem[{Team et~al.(2023)Team, Anil, Borgeaud, Wu, Alayrac, Yu, Soricut, Schalkwyk, Dai, Hauth et~al.}]{team2023gemini}
Gemini Team, Rohan Anil, Sebastian Borgeaud, Yonghui Wu, Jean-Baptiste Alayrac, Jiahui Yu, Radu Soricut, Johan Schalkwyk, Andrew~M Dai, Anja Hauth, et~al. 2023.
\newblock Gemini: a family of highly capable multimodal models.
\newblock \emph{arXiv preprint arXiv:2312.11805}.

\bibitem[{Tong et~al.(2024)Tong, II, Wu, Woo, IYER, Akula, Yang, Yang, Middepogu, Wang, Pan, Fergus, LeCun, and Xie}]{tong2024cambrian}
Shengbang Tong, Ellis L~Brown II, Penghao Wu, Sanghyun Woo, ADITHYA~JAIRAM IYER, Sai~Charitha Akula, Shusheng Yang, Jihan Yang, Manoj Middepogu, Ziteng Wang, Xichen Pan, Rob Fergus, Yann LeCun, and Saining Xie. 2024.
\newblock \href {https://openreview.net/forum?id=Vi8AepAXGy} {Cambrian-1: A fully open, vision-centric exploration of multimodal {LLM}s}.
\newblock In \emph{The Thirty-eighth Annual Conference on Neural Information Processing Systems}.

\bibitem[{Touvron et~al.(2023)Touvron, Martin, Stone, Albert, Almahairi, Babaei, Bashlykov, Batra, Bhargava, Bhosale et~al.}]{touvron2023llama}
Hugo Touvron, Louis Martin, Kevin Stone, Peter Albert, Amjad Almahairi, Yasmine Babaei, Nikolay Bashlykov, Soumya Batra, Prajjwal Bhargava, Shruti Bhosale, et~al. 2023.
\newblock Llama 2: Open foundation and fine-tuned chat models.
\newblock \emph{arXiv preprint arXiv:2307.09288}.

\bibitem[{Wang et~al.(2024)Wang, Bai, Tan, Wang, Fan, Bai, Chen, Liu, Wang, Ge, Fan, Dang, Du, Ren, Men, Liu, Zhou, Zhou, and Lin}]{Qwen2-VL}
Peng Wang, Shuai Bai, Sinan Tan, Shijie Wang, Zhihao Fan, Jinze Bai, Keqin Chen, Xuejing Liu, Jialin Wang, Wenbin Ge, Yang Fan, Kai Dang, Mengfei Du, Xuancheng Ren, Rui Men, Dayiheng Liu, Chang Zhou, Jingren Zhou, and Junyang Lin. 2024.
\newblock Qwen2-vl: Enhancing vision-language model's perception of the world at any resolution.
\newblock \emph{arXiv preprint arXiv:2409.12191}.

\bibitem[{Wei et~al.(2022)Wei, Wang, Schuurmans, Bosma, Xia, Chi, Le, Zhou et~al.}]{wei2022chain}
Jason Wei, Xuezhi Wang, Dale Schuurmans, Maarten Bosma, Fei Xia, Ed~Chi, Quoc~V Le, Denny Zhou, et~al. 2022.
\newblock Chain-of-thought prompting elicits reasoning in large language models.
\newblock \emph{Advances in Neural Information Processing Systems}, 35:24824--24837.

\bibitem[{Wu et~al.(2023)Wu, Yin, Qi, Wang, Tang, and Duan}]{wu2023visual}
Chenfei Wu, Shengming Yin, Weizhen Qi, Xiaodong Wang, Zecheng Tang, and Nan Duan. 2023.
\newblock Visual chatgpt: Talking, drawing and editing with visual foundation models.
\newblock \emph{arXiv preprint arXiv:2303.04671}.

\bibitem[{Wu et~al.(2016)Wu, Schuster, Chen, Le, Norouzi, Macherey, Krikun, Cao, Gao, Macherey et~al.}]{wu2016google}
Yonghui Wu, Mike Schuster, Zhifeng Chen, Quoc~V Le, Mohammad Norouzi, Wolfgang Macherey, Maxim Krikun, Yuan Cao, Qin Gao, Klaus Macherey, et~al. 2016.
\newblock Google's neural machine translation system: Bridging the gap between human and machine translation.
\newblock \emph{arXiv preprint arXiv:1609.08144}.

\bibitem[{Yang et~al.(2024)Yang, Yang, Zhang, Hui, Zheng, Yu, Li, Liu, Huang, Wei et~al.}]{yang2024qwen2}
An~Yang, Baosong Yang, Beichen Zhang, Binyuan Hui, Bo~Zheng, Bowen Yu, Chengyuan Li, Dayiheng Liu, Fei Huang, Haoran Wei, et~al. 2024.
\newblock Qwen2. 5 technical report.
\newblock \emph{arXiv preprint arXiv:2412.15115}.

\bibitem[{Yang et~al.(2023)Yang, Li, Lin, Wang, Lin, Liu, and Wang}]{yang2023dawn}
Zhengyuan Yang, Linjie Li, Kevin Lin, Jianfeng Wang, Chung-Ching Lin, Zicheng Liu, and Lijuan Wang. 2023.
\newblock \href {http://arxiv.org/abs/2309.17421} {The dawn of lmms: Preliminary explorations with gpt-4v(ision)}.
\newblock \emph{arXiv}, abs/2309.17421.

\bibitem[{Yao et~al.(2023)Yao, Jiang, Yang, and Hu}]{yao2023empowering}
Binwei Yao, Ming Jiang, Diyi Yang, and Junjie Hu. 2023.
\newblock Empowering llm-based machine translation with cultural awareness.
\newblock \emph{arXiv preprint arXiv:2305.14328}.

\bibitem[{Zeng et~al.(2023)Zeng, Zhou, Luo, Cheng, and Zhang}]{zeng-etal-2023-cross}
Yan Zeng, Wangchunshu Zhou, Ao~Luo, Ziming Cheng, and Xinsong Zhang. 2023.
\newblock \href {https://doi.org/10.18653/v1/2023.acl-long.315} {Cross-view language modeling: Towards unified cross-lingual cross-modal pre-training}.
\newblock In \emph{Proceedings of the 61st Annual Meeting of the Association for Computational Linguistics (Volume 1: Long Papers)}, pages 5731--5746, Toronto, Canada. Association for Computational Linguistics.

\bibitem[{Zhou et~al.(2022)Zhou, Sch{\"a}rli, Hou, Wei, Scales, Wang, Schuurmans, Cui, Bousquet, Le et~al.}]{zhou2022least}
Denny Zhou, Nathanael Sch{\"a}rli, Le~Hou, Jason Wei, Nathan Scales, Xuezhi Wang, Dale Schuurmans, Claire Cui, Olivier Bousquet, Quoc Le, et~al. 2022.
\newblock Least-to-most prompting enables complex reasoning in large language models.
\newblock \emph{arXiv preprint arXiv:2205.10625}.

\end{thebibliography}

\appendix

\section{Models and Evaluation Protocols}
\label{sec:models}

In this section, we introduce all multimodal models that we evaluate in Section \ref{sec:results}.

\subsection{Open Models}

\subsubsection{Zero-Shot Evaluation (\emph{no labeled data for task})}

\label{subsec:open-models}


Recently, there has been a rise in multimodal language models that are instruction-finetuned to solve tasks in a zero-shot manner \cite{chung2022scaling}. These systems may or may not be trained multilingually. We evaluate these models by providing the models with instructions on solving the task, utilizing the models' zero-shot learning abilities and chain-of-thought reasoning abilities \cite{wei2022chain}. Below, we briefly describe the models that we experiment with under a zero-shot setting:

\paragraph{mBLIP} mBLIP \cite{geigle2023mblip}  extends large multimodal models' capabilities to be multilingual. mBLIP re-align an image encoder previously tuned to an English LLM to a multilingual LLM. Re-alignment training of mBLIP utilizes multilingual data machine-translated from English data.

\paragraph{LLaVA} Large Language and Vision Assistant (LLaVA) is a series of open large multimodal model that are instruction tuned on machine-generated instruction-following data \cite{liu2023llava, liu2023improvedllava, liu2024llavanext}. LLaVA extends the capabilities of existing models by incorporating visual models and large language models. It connects a vision encoder CLIP and an LLM decoder. LLaVA is not explicitly trained to process multilingual data, but the LLM decoder (Vicuna is the default LLM) is known to have seen multilingual data in pretraining \cite{vicuna2023}.

\paragraph{Qwen-VL} Qwen-VL is an open large multilingual multimodal model trained on English and Chinese data. It is based on Qwen-7B \cite{qwen}, incorporating a language-aligned visual encoder and a positionaware adapter. It is trained to be able to process multi-image inputs.

\paragraph{Qwen2-VL} Following Qwen-VL, Qwen2-VL is also trained on English and Chinese data. It is based on Qwen2 \cite{yang2024qwen2}.

\paragraph{Cambrian}
Cambrian-8B is a vision-centric multimodal LLM that focuses on bridging the gap between visual representation learning and language models \cite{tong2024cambrian}. It introduces the Spatial Vision Aggregator (SVA), which efficiently integrates high-resolution visual features with language models. Cambrian also offers a new benchmark called CV-Bench to evaluate 2D and 3D visual understanding. Through its open release of model weights, code, and datasets, Cambrian aims to foster advancements in multimodal AI systems and visual representation research.

\paragraph{Molmo} Trained from scratch, Molmo \cite{deitke2024molmo} is a family of models trained from scratch. It is especially trained on a special 2D-pointing dataset.

\paragraph{Llama3}
Extending Llama 2 \cite{touvron2023llama} with an 8B-parameter model, Llama 3 increases multilinguality, coding, reasoning, and tool usage \cite{llamateam2024llama3herd}. It offers a context length up to 128K, uses grouped-query attention for faster inference, and applies Direct Preference Optimization (DPO) and rejection sampling to align with human preferences, achieving competitive performance across multiple benchmarks.
\subsubsection{Evaluation Post-Finetuning on NLVR2 \emph{(labeled data for task in English)}}
Several end-to-end encoder-based models have been proposed that are pretrained on multilingual multimodal data, and typically need to be fintuned prior to evaluation \cite{devlin2018bert}. Pretraining objectives typically include masked language modeling (text), image-text matching, masked region modeling (image), and multimodal contrastive learning \cite{chen2020uniter, zeng-etal-2023-cross}.

To test on MaRVL, they need to be finetuned on task-specific data. Since MaRVL is a test-only dataset, we finetune on the training data of NLVR2 which is only in English. Note that these models are pretrained on a single image-text pair. To deal with a pair of images in finetuning, each image is separately paired with the statement in two forward passes, and a concatenation of obtained embeddings is passed to a linear classifier to make the prediction. Here, we experiment with CCLM and UNITER-based models as described below. We also finetune mBLIP, but not LLaVa, due to computational constraints introduced by its size. 

\paragraph{UNITER} The UNiversal Image-TExt Representation Learning (UNITERs) framework focuses on achieving end-to-end reasoning across different modalities \cite{chen2020uniter}. This model aims to unify the processing of textual and visual information, fostering more coherent and integrated reasoning capabilities. We experiment with mUNITER and xUNITER, which are initialized from UNITER with mBERT and XLM-R respectively.

\paragraph{CCLM} The Crosslingual Cross-modal Language Model (CCLM) is an open pretrained multilingual multimodal that delves into conditional masked language modeling and contrastive learning techniques to enhance cross-modal understanding \cite{zeng-etal-2023-cross}. This model contribute valuable insights into improving the alignment between textual and visual representations in multilingual scenarios.

\subsection{Proprietary Model GPT-4V}

\paragraph{GPT-4V(ision)} Incorporating multimodality into GPT-4, GPT-4V is able to process image inputs and text inputs together, paving the way for various downstream tasks including visual reasoning tasks \cite{achiam2023gpt, yang2023dawn}. Since GPT-4V is also know for its zero-shot learning abilities \cite{yang2023dawn}, plus finetuning is not supported by GPT-4V\footnote{\url{https://platform.openai.com/docs/guides/fine-tuning/what-models-can-be-fine-tuned}}, we evaluate GPT-4V under a zero-shot setting as discussed in \S\ref{subsec:open-models}.

\paragraph{Gemini-1.5-Pro}
With context length up to one million tokens, Gemini-1.5-Pro \cite{geminiteam2024gemini15unlockingmultimodal} uses a Mixture-of-Experts (MoE) \cite{shazeer2017outrageouslylargeneuralnetworks} design for efficiency. Compared to Gemini 1.0 \cite{geminiteam2024geminifamilyhighlycapable}, it achieves better multimodal reasoning (text, images, video, code), offers in‐context learning, and integrates safety/ethics testing throughout development.

\section{Additional Experiments on Qwen-VL}
\label{sec:additional}

To better understand multilingual and multicultural understanding abilities of our proposed pipeline, we performed additional experiments on Qwen-VL. This is because Qwen-VL is trained on Chinese data, while all other open models we evaluated are pretrained with a focus on English culture, without seeing much data from the local culture. Therefore, in addition to the experiments we discussed in Section \ref{subsec:captioning}, we also performed the third intervention with Qwen-VL on the MaRVL Mandarin Chinese dataset where we caption images using the native language. This experiment resulted in 73.4\% accuracy, while using our interventions with English captions gives 72.8\% accuracy, and using Qwen without interventions gives 58.4\% accuracy. These results extended our points that visio-linguistic models need better understanding of culturally-specific elements. For example, Siheyuan is a culturally specific concept from Chinese culture, where if a model has never seen such concepts previously, it might not be able to generate the correct response for queries containing the concept Siheyuan.

\section{Additional Computation Cost}
\label{sec:cost}
For the first intervention in \S\ref{subsec:translate}, we use the translated statements provided in the MaRVL dataset, so no additional training cost is incurred.

For the second intervention in \S\ref{subsec:visprog}, training cost is not directly comparable, since we finetune ViLT if not using the intervention, and use the pretrained ViLT if using the intervention.

For the third intervention, with a 3\% increase in total evaluation time, we see a 13\% average improvement in performance for LLaVA-v1.5-13B. There is no additional training cost brought by the intervention. Noteworthily, total inference time using LLaVA is halved when using this intervention. 

\section{Machine Translation V.S. Human Translation}

\label{appendix:translation}

In the translation test described in Section \ref{subsec:translate}, we used the Google Translate API \cite{wu2016google}. To investigate whether potential translation inaccuracies could omit certain linguistic nuances in non-English contexts, we also evaluated models on a human-translated version of the dataset.

\begin{table}[ht!]
\centering
\resizebox{\columnwidth}{!}{%
\begin{tabular}{p{0.25\columnwidth} p{0.35\columnwidth} p{0.35\columnwidth}}
\toprule
\textbf{Model} & \textbf{Machine (zh)} & \textbf{Human (zh)}\\
\midrule
xUNITER & 63.3 & 64.4\\
GPT-4V  & 78.4 & 79.9\\
\bottomrule
\end{tabular}%
}
\caption{MaRVL-ZH results (Machine translation vs.\ Human translation).}
\label{tab:human-translation}
\end{table}

Specifically, we tested xUNITER (finetuned on NLVR2) and GPT-4V (zero-shot) on the Chinese subset of MaRVL using human-translated data provided by the original MaRVL paper \cite{liu-etal-2021-visually}. Table~\ref{tab:human-translation} shows the results. Notably, the human-translated data yields only marginal improvements over machine translation. While we acknowledge the limitations inherent in the translation-based approach, these findings support using machine translation for broader evaluations due to its practicality under resource constraints.

Moreover, the small discrepancy in performance between human and machine translations suggests that the translation method itself may have only minor influences on model performance. Accordingly, we relied on the Google Translate API, consistent with the MaRVL translate test setting \cite{liu-etal-2021-visually}.
\end{document}